\documentclass[conference]{IEEEtran}
\IEEEoverridecommandlockouts
\usepackage{cite}
\usepackage{amsmath,amssymb,amsfonts}
\usepackage{algorithmic}
\usepackage{graphicx}
\usepackage{textcomp}
\usepackage{xcolor}
\usepackage{booktabs}
\usepackage{multirow}
\usepackage{bm}
\usepackage{array}
\usepackage[normalem]{ulem}
\usepackage[colorlinks,urlcolor=blue]{hyperref}
\def\BibTeX{{\rm B\kern-.05em{\sc i\kern-.025em b}\kern-.08em
    T\kern-.1667em\lower.7ex\hbox{E}\kern-.125emX}}

\def\0{{\mathbf 0}}
\def\1{{\mathbf 1}}

\def\c{{\mathbf c}}
\def\d{{\mathbf d}}

\def\g{{\mathbf g}}
\def\h{{\mathbf h}}

\def\o{{\mathbf o}}

\def\s{{\mathbf s}}
\def\t{{\mathbf t}}

\def\x{{\mathbf x}}
\def\y{{\mathbf y}}

\def\I{{\mathbf I}}

\def\ie{{\textit{i.e.}}}

\def\0{{\mathbf 0}}
\def\1{{\mathbf 1}}

\def\c{{\mathbf c}}
\def\d{{\mathbf d}}

\def\g{{\mathbf g}}
\def\h{{\mathbf h}}

\def\o{{\mathbf o}}

\def\s{{\mathbf s}}
\def\t{{\mathbf t}}

\def\x{{\mathbf x}}
\def\y{{\mathbf y}}

\def\I{{\mathbf I}}

\def\ie{{\textit{i.e.}}}

\begin{document}

\title{Towards Robust Time-of-Flight Depth Denoising with Confidence-Aware Diffusion Model}

\author{\IEEEauthorblockN{Changyong He}
\IEEEauthorblockA{\textit{School of Computer Sciene and} \\
\textit{Technology, Tongji University}\\
Shanghai, China \\
2433293@tongji.edu.cn}
\and
\IEEEauthorblockN{Jin Zeng$^{\ast}$ \thanks{* Corresponding author}}
\IEEEauthorblockA{\textit{School of Computer Sciene and} \\
\textit{Technology, Tongji University}\\
Shanghai, China \\
zengjin@tongji.edu.cn}
\and
\IEEEauthorblockN{Jiawei Zhang}
\IEEEauthorblockA{
\textit{SenseTime Research}\\
Shenzhen, China \\
zhjw1988@gmail.com}
\and
\IEEEauthorblockN{Jiajie Guo}
\IEEEauthorblockA{\textit{School of Computer Sciene and} \\
\textit{Technology, Tongji University}\\
Shanghai, China \\
2433292@tongji.edu.cn}

\thanks{The work was supported in part by National Natural Science Foundation of China under Grant 62201389 and in part by Shanghai Rising-Star Program under Grant 22YF1451200.}
}


\maketitle
\vspace{-3cm}
\begin{abstract}
Time-of-Flight (ToF) sensors efficiently capture scene depth, but the nonlinear depth construction procedure often results in extremely large noise variance or even invalid areas.
Recent methods based on deep neural networks (DNNs) achieve enhanced ToF denoising accuracy but tend to struggle when presented with severe noise corruption due to limited prior knowledge of ToF data distribution.
In this paper, we propose DepthCAD, a novel ToF denoising approach that ensures global structural smoothness by leveraging the rich prior knowledge in Stable Diffusion and maintains local metric accuracy by steering the diffusion process with confidence guidance.
To adopt the pretrained image diffusion model to ToF depth denoising, we apply the diffusion on raw ToF correlation measurements with dynamic range normalization before converting to depth maps.
Experimental results validate the state-of-the-art performance of the proposed scheme, and the evaluation on real data further verifies its robustness against real-world ToF noise.
\end{abstract}


\section{Introduction}
\label{sec:intro}
With the capacity to produce high-resolution depth maps using hardware with low power consumption, Time-of-Flight (ToF) depth sensing has become a cornerstone technique for 3D scene capture \cite{zanuttigh2016time}. Among the various types of ToF sensors, continuous-wave ToF sensors stand out for their low cost and real-time capabilities, making them widely used in applications such as 3D reconstruction \cite{zhang2015fusion}, scene understanding \cite{zeng2019deep}, robotics \cite{gao2021joint}, etc. 
For brevity, we hereinafter refer to continuous-wave Time-of-Flight sensors as ToF sensors.
However, depth images captured by commercial ToF sensors such as Microsoft Kinect suffer from severe noise on dark and distant
surfaces, leading to inaccurate depth maps \cite{kurillo2022evaluating}.
This motivates various ToF depth restoration methods to enhance depth imaging quality \cite{xiang2016libfreenect2,chen2020very,deng2023two,su2018deep,uda2019,schelling2022radu,jia2025deep}.

\begin{figure}[t]
    \centering
    \includegraphics[width=1.0\linewidth]{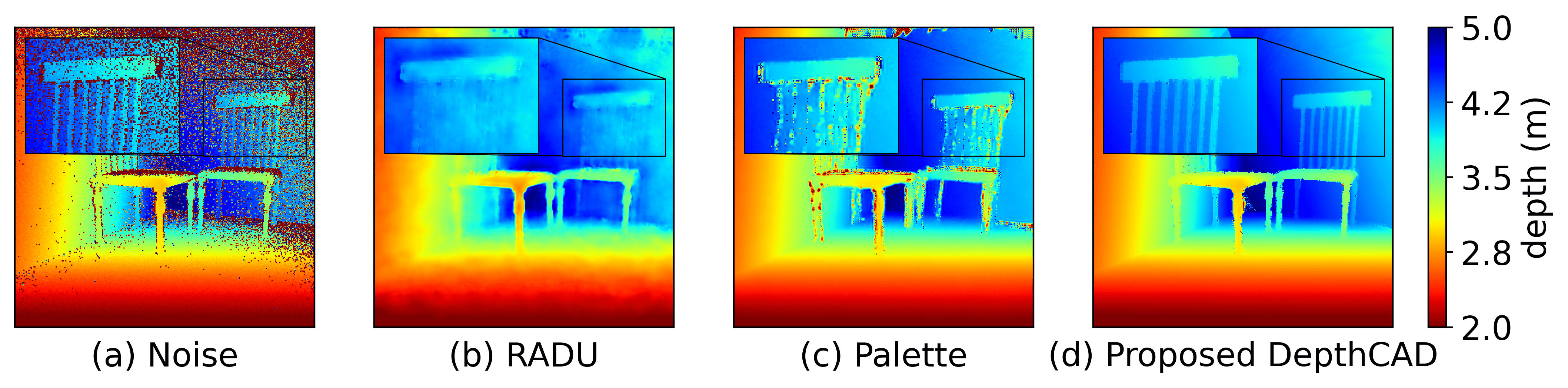}
    \vspace{-0.7cm}
    \caption{Results of ToF depth denoising with example from FLAT dataset \cite{guo2018tackling}: (a) noisy input with high dynamic range of noise variance, results with (b) DNN-based RADU \cite{schelling2022radu}, (c) diffusion-based Palette \cite{saharia2022palette} and (d) proposed DepthCAD. As highlighted in black rectangles, RADU exhibits smooth but blurry details and Palette exhibits inaccurate depth estimation, while DepthCAD generates accurate depth with detail preservation.}
    \label{fig:denoise_failure}
    \vspace{-0.3cm}
\end{figure}

Model-based methods are based on mathematical models for depth images, \textit{e.g.}, bilateral filter \cite{barron2016fast}, non-local means \cite{georgiev2018time}, 3D block-matching \cite{dabov2007image}, graph-based image priors \cite{hu2013depth,rossi2020joint}, \textit{etc.}, but the assumed models are usually constructed based on hand-crafted features, leading to inaccurate optimization formulations and sub-par performance. 
More recent approaches utilize end-to-end deep neural networks (DNNs) for ToF depth denoising and achieve state-of-the-art (SOTA) performance \cite{su2018deep,uda2019,schelling2022radu,dong2025exploiting}.
For example, ToFNet \cite{su2018deep} directly processes raw ToF correlation measurements to produce depth maps with an end-to-end deep learning framework, which eliminates cumulative errors and information loss found in traditional multi-step pipelines, significantly improving the accuracy of the depth maps.
However, because of the nonlinear depth construction process in ToF imaging, the noise distribution is usually nonuniform, \textit{i.e.}, the signal-to-noise ratio (SNR) has a large dynamic range across the whole scene as shown in Fig.~\ref{fig:denoise_failure}(a). 
For example, in background regions with particularly high noise, the captured data becomes invalid, making denoising extremely challenging as illustrated in Fig.~\ref{fig:denoise_failure}(b) where DNN-based method RADU \cite{schelling2022radu} shows blurry results in high-confidence areas and fails to effectively reconstruct invalid regions due to \textit{limited prior knowledge about the underlying scene structure and real noise characteristics.}

To tackle the above problems, we derive the ToF denoising network from Stable Diffusion by utilizing the rich prior knowledge encapsulated in the diffusion model to capture the geometric scene structure.
Despite the high noise levels or invalid regions in the corrupted inputs, the model is able to generate depth maps with global structural smoothness.


While diffusion models generate expressive results in color image denoising \cite{ho2020denoising,wang2022zero}, these models are typically designed for image generation tasks and can hardly generalize to depth domain. 
For example, when applied to depth denoising, diffusion-based Palette \cite{saharia2022palette} results in inaccurate depth estimation shown in Fig.~\ref{fig:denoise_failure}(c).
These challenges arise from two factors.
1) These diffusion models are primarily trained on color images and are limited in generalizing the prior knowledge of color images to depth maps.
\textit{This domain disparity between RGB and depth data} hinders the diffusion model to unveil the underlying spatial relationships and recover precise depth values.
2) 
\textit{Without effective fidelity preservation} of confident input values, the diffusion models struggle to balance local fidelity and generative quality, 
which undermines the depth recovery accuracy. 
While methods like StableSR \cite{wang2024exploiting} attempt to enhance fidelity by employing a global coeffecient to control the balance between generative refinement and fidelity preservation, they often produce inaccurate results due to the absence of confidence information to distinguish specific regions that demand fidelity maintenance from those require generative refinement.

In contrast, we propose \underline{\textbf{Depth}} Denoising with \underline{\textbf{C}}onfidence-\underline{\textbf{A}}ware \underline{\textbf{D}}iffusion model (DepthCAD) to address the above limitations.
Specifically, based on the analysis of ToF depth image degradation process, we apply Stable Diffusion on the raw correlation measurements captured by ToF sensors before converted to depth for denoising, as their noise distribution is more similar to that of color images than that of depth maps.
Furthermore, to compensate for the distribution difference between raw correlations and color images, we apply nonlinear dynamic range normalization to the correlation data, effectively alleviating domain disparity and bypassing the need for specialized depth diffusion models. 
Additionally, we compute confidence based on the noise levels of the noisy depth map, which is integrated into the diffusion to distinguish high-confidence areas prioritizing fidelity preservation, and less reliable areas focusing on generative refinement.
In summary, our main contributions are as follows:
\begin{itemize}
    \item We derive the ToF depth denoising network from the diffusion model by leveraging its powerful prior knowledge to achieve enhanced global structural smoothness in resulting depth maps; 
    \item We adapt the image diffusion model to ToF depth denoising by performing diffusion on raw correlations with dynamic range normalization to bridge the domain gap between ToF depth and color images;
    \item We design confidence-aware guidance to regulate the diffusion process toward accurate depth restoration, mitigating the generative bias of the diffusion model to enhance metric accuracy. 
\end{itemize}
Experimental results validate our SOTA performance and demonstrate the robustness of DepthCAD across both synthetic and real-world ToF datasets, achieving superior denoising accuracy and fidelity preservation.
Code is available at \href{https://github.com/BadbeardHe/DepthCAD}{\textit{https://github.com/BadbeardHe/DepthCAD}}.

\section{ToF Noise Analysis}
\label{sec:tof}
To measure the depth $d$ of an object, the ToF camera emits a periodic signal $s_e(t)$ modulated by a sinusoidal function with frequency $f_m$ and receives the reflected signal $s_r(t)$: 
\begin{equation}
    s_e(t) = s_0 (1+ \cos(2\pi f_m t)),  s_r(t) = \alpha \cos(2\pi f_m t - \phi) + \beta,
\end{equation}
where $s_0$ and $\alpha$ are the signal amplitudes, $\beta$ is the ambient light intensity, and $\phi$ denotes the phase shift after the signal has traveled the distance $2d$ \cite{zanuttigh2016time}. Without loss of generality, we assume $s_0 = 1$.
The phase shift $\phi$ is then measured by the correlation between $s_r(t)$ and the phase shifted version of $s_e(t)$ with phase offset $\theta$, resulting in raw measurements \cite{zanuttigh2016time}
\begin{align}
c_{\theta} & = \lim_{T \to \infty} \frac{1}{T} \int_{-\frac{T}{2}}^{\frac{T}{2}} s_r(t) s_e(t+ \frac{\theta}{2\pi f_m }) dt = \frac{\alpha}{2} \cos(\phi+\theta) + \beta. \label{eq:amp}
\end{align}
By measuring $c_{\theta}$ for multiple phase offsets $\theta$, the \textit{correlation measurements}, \textit{i.e.}, the in-phase $i$ and quadrature $q$ components of $\phi$ are computed as \cite{schelling2022radu},
\begin{equation} \label{eq:iq}
    x_i = \sum_{\theta} \cos(\theta) c_{\theta}, ~ x_q = \sum_{\theta} - \sin(\theta) c_{\theta},
\end{equation}
Given $\phi = \arctan (x_q/x_i)$, the distance is constructed as \cite{schelling2022radu},
\begin{equation}\label{eq:raw2d}
    x_d = \frac{c  }{4 \pi f_m}  \arctan (x_q/x_i),
\end{equation}
where $c$ is the light speed. 

To investigate how noise in the raw correlations $x_i$ and $x_q$ affects the depth estimation, it is commonly assumed that the noisy versions of $x_i$ and $x_q$, \ie, $y_i$, $y_q$, are independent and identically distributed with bivariate Gaussian distribution \cite{frank2009theoretical,georgiev2018time}, 
\begin{equation} \label{eq:pdf_iq}
P(y_i,y_q|x_i,x_q) = \frac{1}{2 \pi \sigma^2} \exp (-\frac{(x_i-y_i)^2 + (x_q-y_q)^2}{2 \sigma^2}),
\end{equation}
where $\sigma$ is the noise variance.
Under normal noise level, \ie, $\gamma=\sigma/y_a \ll 1$, where $y_a$ is the noisy amplitude, the distribution of depth noise $n_d$ is derived in \cite{georgiev2018time} as
\begin{equation} \label{eq:pdf_d_simple}
 P(n_d) \approx \frac{\cos (4\pi f_m n_d/c)}{\gamma \sqrt{2\pi}} \exp ( \, -\frac{\sin^2(4\pi f_m n_d/c)}{2\gamma^2} ) \,.
\end{equation} 

From (\ref{eq:pdf_d_simple}), it is evident that the nonlinear depth construction procedure (\ref{eq:raw2d}) leads to the noise distribution of depth close to a Gaussian distribution with variance approximately proportional to $\gamma /f_m$. This indicates that the noise variance has a large dynamic range and can become substantial in case of 
very low signal amplitude $y_a$, posing significant challenges for denoising algorithms.
This motivates the design of our confidence-aware diffusion model DepthCAD. 
By utilizing raw correlations instead of depth maps as input, we mitigate the domain gap between ToF data and color images for which the production-ready diffusion models are optimized. 
Additionally, we incorporate confidence according to the noise levels of corrupted depth maps into the diffusion process to ensure effective fidelity preservation.

\section{Proposed DepthCAD Model}
\label{sec:algo}
In this section, we present the design details of the proposed DepthCAD, a confidence-aware diffusion model for ToF depth denoising that generates globally smooth depth estimation while maintaining local fidelity preservation.

\begin{figure*}[t]
    \centering
    \includegraphics[width=1.0\linewidth]{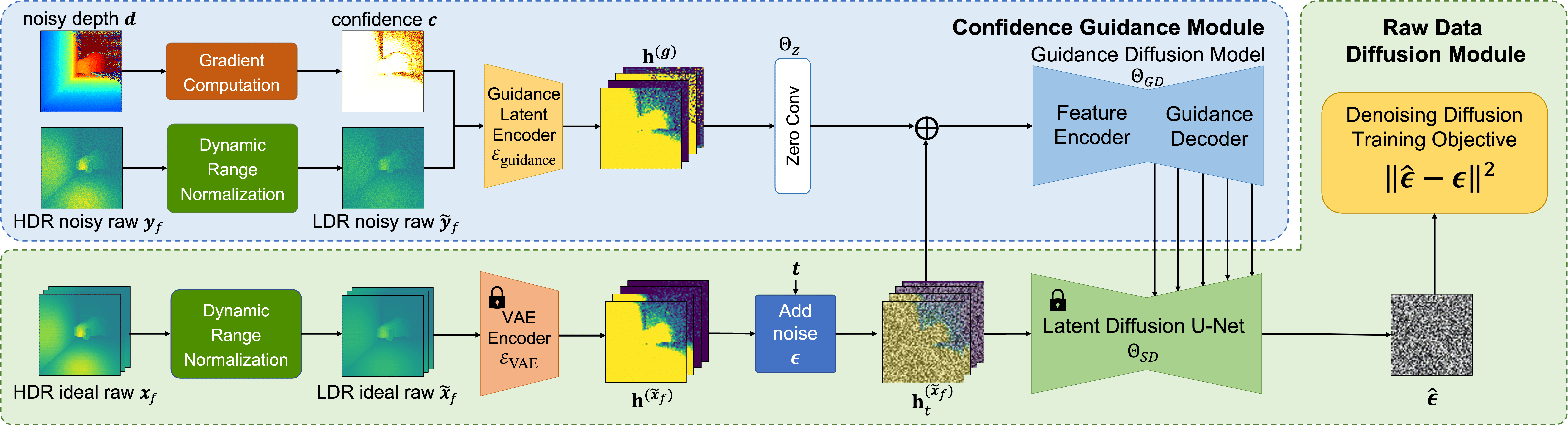}
    \vspace{-0.7cm}
    \caption{Overview of the training process. 
    We derive the Raw Data Diffusion Module from pretrained Stable Diffusion 2.1, performing diffusion on raw correlation $\tilde{\x}_f$ with dynamic range normalization to bridge the domain gap between ToF depth and color images, which enables the rich prior knowledge of Stable Diffusion for enhanced global structural smoothness.
    To balance generative quality and fidelity preservation, we design the Confidence Guidance Module with confidence $\c$ from gradient computation of noisy depth $\d$ and normalized noisy raw correlation $\tilde{\y}_f$ as guidance. These guidance are fused with the inputs to exert influence on the diffusion process via connections from Guidance Diffusion Model to Latent Diffusion U-Net.
    During training, all components in the Raw Data Diffusion Module are frozen, we train the Confidence Guidance Module by optimizing the standard diffusion objective.
    }
    \label{fig:training}
    \vspace{-0.3cm}
\end{figure*}

\subsection{Overall Framework}
As shown in Fig.~\ref{fig:training}, DepthCAD consists of two key components: the Raw Data Diffusion Module and the Confidence Guidance Module. The Raw Data Diffusion Module leverages Stable Diffusion \cite{rombach2022high} for its powerful prior knowledge to enable reasonable depth reconstruction. The Confidence Guidance Module incorporates noisy raw correlations and corresponding confidence information into the diffusion process to ensure global generative smoothness while preserving local fidelity.

\subsection{Raw Data Diffusion Module}
\begin{figure}[t]
    \centering
    \includegraphics[width=1.0\linewidth]{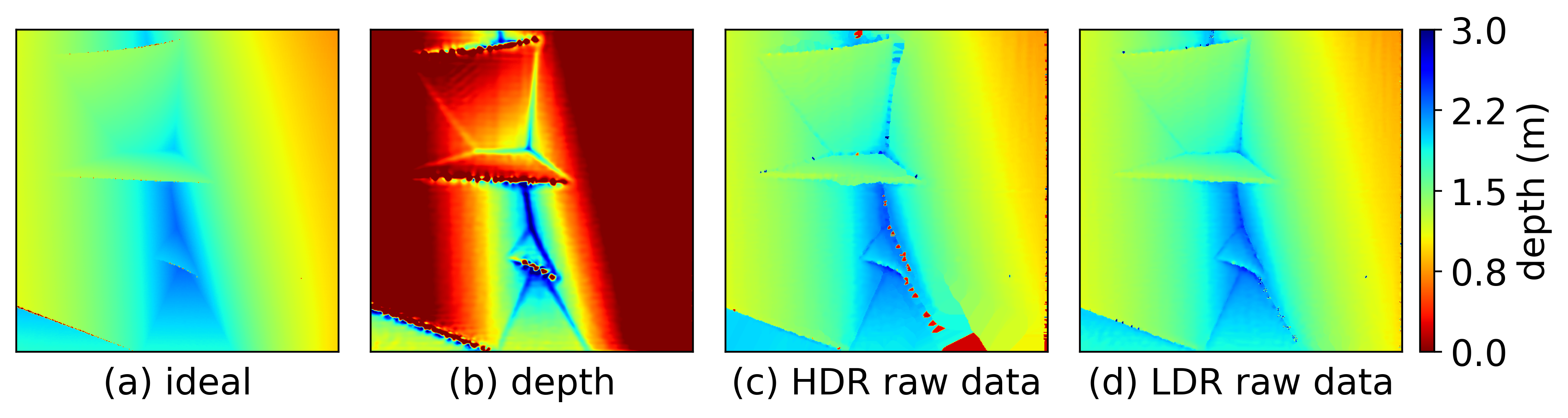}
    \vspace{-0.8cm}
    \caption{Comparison of reconstruction results using the Stable Diffusion VAE with different input formats. (a) ideal depth map,  reconstruction results from (b) depth measurements, (c) unnormalized raw correlations and (d) normalized raw correlations.}
    \label{fig:cmp_vae}
    \vspace{-0.4cm}
\end{figure}
We use Stable Diffusion 2.1 \cite{rombach2022high} as backbone in the Raw Data Diffusion Module to enhance the generative process for ToF denoising. As a latent diffusion model (LDM), it operates in a compact latent space, which reduces computational costs and enables the model to focus on relevant features while preserving depth-specific structures.

Based on the analysis of ToF noise (\ref{eq:pdf_iq}) and (\ref{eq:pdf_d_simple}), the noise distribution of raw correlations is closer to the RGB input typically used by Stable Diffusion than depth data. Therefore, instead of directly denoising depth measurements, our method integrates the denoising process into the ToF imaging pipeline before depth construction and denoise the raw correlations in (\ref{eq:iq}).
However, the raw correlations exhibit an extremely large dynamic range, sometimes with values varying by over a thousandfold.
To further reduce the distribution difference between raw correlations and color images, we apply dynamic range normalization. 
Specifically, for a pair of raw correlations $(\x_i, \x_q)$ under a certain frequency, the dynamic range is first compressed using: 
\begin{equation}
R = |\x_i| + |\x_q|, ~R_{\text{LDR}} = 16 \sqrt{R + 36} - 96.
\end{equation}
The raw correlations are then normalized as:
\begin{equation}
\tilde{\x}_i = \frac{R_{\text{LDR}} \cdot \x_i}{R}, ~\tilde{\x}_q = \frac{R_{\text{LDR}} \cdot \x_q}{R}.
\end{equation}
To address cases where $R = 0$ (\textit{e.g.}, regions with no signal), we substitute $R = 1$ to avoid division by zero to ensure stability.

In this way, in normal range where most measurements fall in, the normalized values remain mostly intact. For measurements in relatively high range, \textit{e.g.}, in glossy areas, the values are greatly compressed.
After this transformation, the raw correlations are scaled to a low-dynamic-range (LDR) format within the value range $[-1, 1]$, which matches the input requirement of the variational autoencoder (VAE) of Stable Diffusion.

To validate the effectiveness of the dynamic range normalization, we compare the reconstruction quality of different inputs using the Stable Diffusion VAE. 
As shown in Fig.~\ref{fig:cmp_vae}, the reconstruction on depth map fails due to its dissimilarity to color images, and unnormalized raw correlation measurements leads to artifacts caused by abnormal values during the depth construction as in (\ref{eq:raw2d}) due to excessive dynamic ranges. In contrast, normalized raw correlations achieve superior reconstruction quality, preserving structural details and ensuring stability. 
These results highlight the importance of dynamic range normalization in reducing the distribution difference between raw correlations and color images for effective denoising, enabling the utilization of the rich prior knowledge to effectively model the intricate noise patterns and characteristics of raw correlations in the diffusion module.

During training, we freeze the weights in the Raw Data Diffusion Module.
This ensures that the pre-trained generative priors remain intact while the Confidence Guidance Module adapts to the specific noise and structural characteristics of raw correlations. 

\vspace{-0.2cm}
\subsection{Confidence Guidance Module}
While we perform denoising on raw correlations, the ultimate objective is to construct a clean and accurate depth map using denoised raw correlations through (\ref{eq:raw2d}). To ensure more targeted and effective guidance, we calculate the confidence based on the gradient variations of the noisy depth map, which reflects its noise levels and measurement reliability. 
Specifically, for a given noisy depth map $d(u, v)$, the gradients along the horizontal and vertical directions are computed using the Sobel operator, and the gradient magnitude $d_\text{mag}$ at each pixel is given by:
\begin{equation}
d_u = \frac{\partial d}{\partial u},~d_v = \frac{\partial d}{\partial v},~
d_\text{mag} = \sqrt{d_u^2 + d_v^2}.
\end{equation}
To ensure consistency across varying depth values, the gradient magnitude is normalized to the range $[0, 1]$, and the confidence map $c$ is then defined as:
\begin{equation}
 \tilde{d}_\text{mag} = \frac{d_\text{mag} - \min(d_\text{mag})}{\max(d_\text{mag}) - \min(d_\text{mag})},~c = 1 - \tilde{d}_\text{mag}.
\end{equation}

To effectively guide the diffusion process toward generating accurate depth with high fidelity and sharp details, we introduce the Confidence Guidance Module. This module leverages normalized noisy raw correlation $\tilde{\y}_f$ for structural information and confidence map $\c$ to encode measurement reliability, which provide complementary guidance to balance the quality of generation and fidelity to the original scene.
Specifically, we apply a zero convolution on the guidance input, which is then fused with the latent input from the Raw Data Diffusion Module.
Next, we create a trainable copy named Guidance Diffusion Model from the frozen Stable Diffusion and replace the decoder with a guidance decoder. The Guidance Diffusion Model takes the fused features as input and the outputs are integrated to the corresponding decoder blocks of the Raw Data Diffusion Module via:
\begin{equation}
     \h^{(\o)} = \mathcal{F}_{\Theta_{\textit{SD}}}(\h^{(\tilde{\x}_f)}) + \mathcal{F}_{\Theta_\textit{GD}}(\mathcal{F}_{\Theta_z}(\h^{(\g)}) + \h^{(\tilde{\x}_f)}),
\end{equation}
where $\h^{(\o)}$ is the output of the decoder block in Latent Diffusion U-Net, $\mathcal{F}_{\Theta_{\textit{SD}}}$, $\mathcal{F}_{\Theta_\textit{GD}}$ and $\mathcal{F}_{\Theta_z}$ represent the operation of Stable Diffusion, the Guidance Diffusion Model, and zero convolution, respectively. $\h^{(\g)}$ and $\h^{(\tilde{\x}_f)}$ are the latent representations of guidance and the normalized ideal raw correlations. 
In this way, the guidance can exert meaningful influence on the diffusion process. 


\subsection{Training}
\label{subsec:training}
The overall training pipeline is illustrated in Fig.~\ref{fig:training}. 
Since the VAE from Stable Diffusion is designed for 3-channel (RGB) inputs, given a normalized single-channel ideal raw correlation $\tilde{\x}_f$ under frequency $f$, we replicate the data into three channels. 
The normalized ideal raw correlation $\tilde{\x}_f$, the confidence guidance $\c$, and the normalized noisy raw correlation guidance $\tilde{\y}_f$ are encoded into the latent space by
\begin{equation}
\h^{(\tilde{\x}_f)}=\mathcal{E}_{\text{VAE}}(\tilde{\x}_f),~\h^{(\g)}=\mathcal{E}_\textit{guidance}(\text{concat}(\tilde{\y}_f, \c)),
\end{equation}
where $\mathcal{E}_{\text{VAE}}$ represents the encoder of the frozen VAE, and $\mathcal{E}_\textit{guidance}$ represents the guidance latent encoder in the Confidence Guidance Module. 
Additionally, the model is conditioned on timestep $\t$ and text prompt $\s$. 
Following the training procedure in Stable Diffusion \cite{rombach2022high}, 
the diffusion algorithm learns a network $f_\theta$ to predict the noise added to the noisy correlation $\h^{(\tilde{\x}_f)}_t$ by minimizing
\begin{equation}
\mathcal{L}=\mathbb{E}_{\h^{(\tilde{\x}_f)}_0, \h^{(\g)}, \t, \s, \bm{\epsilon} \sim \mathcal{N}(0,\I)}\left[\| \bm{\epsilon}-f_\theta\left(\h^{(\tilde{\x}_f)}_0, \h^{(\g)}, \t, \s \right) \|_2^2\right].
\end{equation}

Stable Diffusion was primarily trained for image generation conditioned on text prompts. However, in our approach, the text prompt input is set to an empty string. This prompt-free configuration ensures that the model focuses entirely on the structural patterns and noise characteristics of the raw correlations, avoiding extraneous influence from text prompts. 

\subsection{Inference}
During inference, the latent raw correlations is initialized as Gaussian noise. The trained model takes the noisy raw correlations and confidence maps as guidance to steer the diffusion process, ultimately producing a denoised and refined raw correlations with global smoothness and detail sharpness. As in the training process, the text prompt is set to an empty string, ensuring that the model relies entirely on the data distribution and learned priors. 
The output is decoded via the VAE decoder, then averaged across channels, and further processed through (\ref{eq:raw2d}) to generate a clean depth map.

\section{Experiments}
\label{sec:exp}
\begin{figure*}[th]
    \centering
    \includegraphics[width=1.0\textwidth]{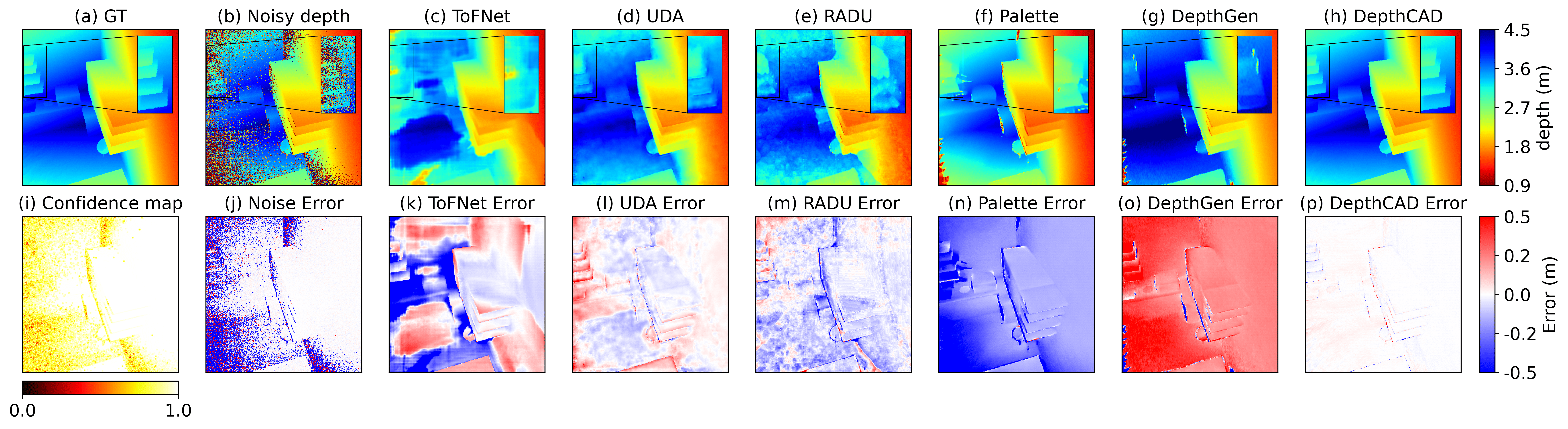}
    \vspace{-0.8cm}
    \caption{Depth results and error maps of ToF depth denoising on FLAT dataset \cite{guo2018tackling}: (a) GT, (b) noisy depth, results of (c) ToFNet \cite{su2018deep}, (d) UDA \cite{uda2019}, (e) RADU \cite{schelling2022radu}, (f) Palette \cite{saharia2022palette}, (g) DepthGen \cite{saxena2023monocular}, and (h) proposed DepthCAD. Corresponding error maps are in the second row. (i) shows the confidence map we use as guidance condition.}
    \label{fig:flat_cmp}
    \vspace{-0.3cm}
\end{figure*}

\begin{figure*}[th]
    \centering
    \includegraphics[width=1.0\textwidth]{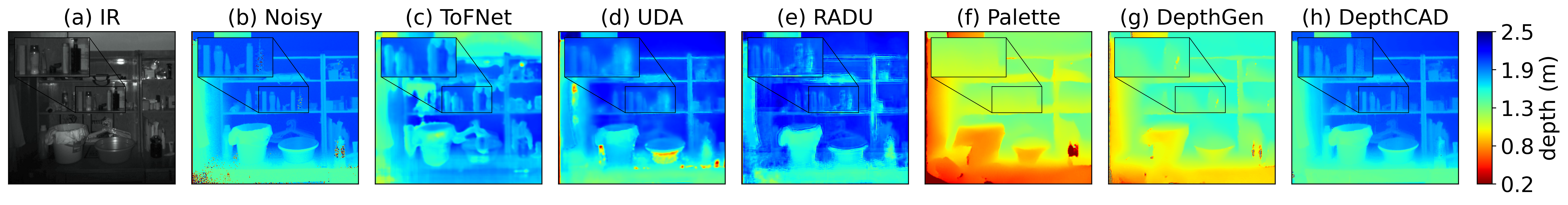}
    \vspace{-0.9cm}
    \caption{Visual results of ToF depth denoising on real data captured by Kinect v2 sensor: (a) infrared image, (b) noisy depth captured by Kinect v2, and results of (c) ToFNet \cite{su2018deep}, (d) UDA \cite{uda2019}, (e) RADU \cite{schelling2022radu}, (f) Palette \cite{saharia2022palette}, (g) DepthGen \cite{saxena2023monocular}, and (h) proposed DepthCAD.}
    \label{fig:real_cmp}
    \vspace{-0.2cm}
\end{figure*}

In this section, we experimentally verify the contribution of the proposed DepthCAD for ToF data denoising through comparison with a range of baseline and state-of-the-art methods on a synthetic dataset FLAT \cite{guo2018tackling} and on real-world Kinect v2 data. We further validate the effectiveness of individual components within our architecture via ablation study.

\subsection{Experimental Setup}
\textbf{Dataset.}
We use the synthetic FLAT dataset \cite{guo2018tackling} for training and testing, which consists of 1923 depth images and corresponding raw correlations with simulated Kinect noise. From this, 1815 sets of raw correlations are used for training, and 108 sets for testing. To evaluate the generalization ability on real-world ToF data, we further collect data using a Kinect v2.

\textbf{Training details.}
We implement DepthCAD using PyTorch framework\cite{paszke2017automatic}. 
For training, we use the DDPM noise scheduler \cite{ho2020denoising} with 1000 sampling steps. We accumulate gradients over 4 steps and employ an 8-bit optimizer with a learning rate of $1 \times 10^{-5}$ to optimize memory usage. The model is trained for 20K iterations with batch size of 16 on a single NVIDIA RTX 3090. 
During inference, we use DDIM scheduler \cite{song2020denoising} with 20 sampling steps.

\textbf{Metrics.}
We evaluate the performance of our method using five widely adopted metrics: Mean Absolute Error (MAE), Absolute Relative Error (AbsRel), and $\delta_1$, $\delta_2$ and $\delta_3$ accuracy following \cite{ke2024repurposing}. Lower values of MAE and AbsRel and larger $\delta_1$, $\delta_2$ and $\delta_3$ indicate higher accuracy.

\subsection{Comparison with Existing Schemes}
We compare our method with model-based approach libfreenect2 \cite{xiang2016libfreenect2}, BM3D \cite{dabov2007image}, DNN-based approaches ToFNet \cite{su2018deep}, UDA \cite{uda2019}, RADU \cite{schelling2022radu}, and diffusion-based approaches Palette \cite{saharia2022palette}, DepthGen \cite{saxena2023monocular}. To ensure a fair comparison, all competing methods are retrained and tested on the same FLAT dataset and evaluation metrics. In addition, following \cite{barron2016fast}, we augment FLAT dataset with simulated edge noise. Note that the same model is used for testing in both noise settings to test generalization ability to unseen noise. As shown in Table.~\ref{tab:comparison}, DepthCAD outperforms  other methods in both noise settings. 

For visual evaluation, we present qualitative comparison in Fig.~\ref{fig:flat_cmp}. DepthCAD accurately reconstructs the overall structure of the scene and preserves fine details, as highlighted in the zoomed-in region, further confirming DepthCAD’s ability to maintain high fidelity while effectively removing noise.

It is worth noting that diffusion-based methods Palette \cite{saharia2022palette} and DepthGen \cite{saxena2023monocular} exhibit suboptimal performance due to the domain gap between color images and ToF depth maps, as they are primarily designed for tasks like image generation and struggle when directly applied to depth denoising. 

\vspace{-0.1cm}
\subsection{Ablation Study}
To further investigate the impact of each component in DepthCAD, we conduct a series of ablation experiments on FLAT dataset \cite{guo2018tackling} with different variants of DepthCAD. 
Quantitative results in Table~\ref{tab:ablation} validate that each component is essential for denoising accuracy and stability.

\textbf{Confidence Guidance Module.}
We remove the Confidence Guidance Module and use the pre-trained StableDiffusionImg2ImgPipeline \cite{rombach2022high} to infer depth from raw correlations. 
Without structural cues, the model operate entirely on the data distribution learned during pre-training, leading to tremendous performance drop. This confirms that the Confidence Guidance Module is essential for directing the diffusion process towards more stable and reasonable outputs.

\textbf{Guidance condition of the noisy raw data.}
Without direct noisy raw measurements, the model has to infer from confidence maps, which provide only an indication of measurement reliability. This results in significant performance degradation across all metrics, highlighting the importance of the noisy raw correlations for explicit fidelity information. We further evaluate the impact of dynamic range normalization by incorporating high-dynamic-range (HDR) noisy raw correlations as guidance. The noticeable performance decline confirms the effectiveness of dynamic range normalization in bridging the domain gap between raw correlations and color images.

\begin{figure}[t]
    \centering
    \includegraphics[width=1.0\linewidth]{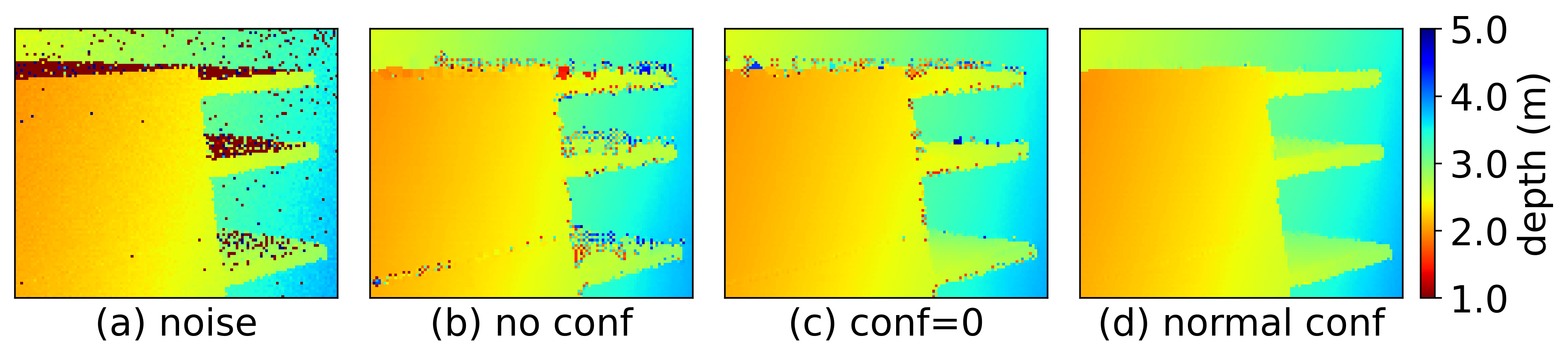}
    \vspace{-0.8cm}
    \caption{Comparison of denoising results under different confidence settings. (a) Noisy depth input, (b) depth generated with: (b) no confidence, (c) confidence uniformly set to 1, and (d) normal confidence information.}
    \label{fig:cmp_conf}
    \vspace{-0.4cm}
\end{figure}
\textbf{Guidance condition of the confidence map.}
When using only the noisy raw correlations as guidance, the model performance declines, as the lack of confidence guidance hinders fidelity preservation. 
To further evaluate the role of confidence, we set the confidence map entirely to zero. As illustrated in Fig.~\ref{fig:cmp_conf}(c), the result exhibits increased artifacts and instability in the output,  highlighting the importance of confidence in balancing generation quality and fidelity preservation.

\textbf{Raw Denoising Module.}
We replace the Stable Diffusion 2.1 backbone with the earlier version Stable Diffusion 1.5, whose learned priors are less refined. This substitution leads to a slight performance drop, indicating that the improved priors in SD2.1 enhance depth reconstruction. Nevertheless, with the noisy raw correlations and confidence as guidance, the degraded SD1.5 is still able to generate reasonable results.

\subsection{Generalization Evaluation with Real Kinect v2 Data}
To assess the generalization ability of DepthCAD on real-world data, we capture ToF data with Kinect v2 sensor and conduct qualitative comparison shown in Fig.~\ref{fig:real_cmp}. 
DNN-based methods ToFNet \cite{su2018deep}, UDA \cite{uda2019}, and RADU \cite{schelling2022radu} generate over-smoothed and blurred results, and diffusion-based approaches Palette \cite{saharia2022palette} and DepthGen \cite{saxena2023monocular} exhibit noticeable depth offsets. In contrast, DepthCAD, despite being trained on synthetic data, shows great performance on real-world measurements. This highlights its strong generalization capability for practical ToF depth denoising.

\begin{table}[t]
\centering
\caption{Comparison of denoising accuracy on FLAT testing dataset and augmented dataset}
\vspace{-0.2cm}
\resizebox{\columnwidth}{!}{ 
\begin{tabular}{lccccccc}
\toprule
\multirow{2}{*}{Methods} & \multicolumn{3}{c}{FLAT Dataset} & \multicolumn{3}{c}{FLAT Dataset with augmented noise} \\
\cmidrule(lr){2-4} \cmidrule(lr){5-7}
 & MAE(m)$\downarrow$ & AbsRel$\downarrow$ & $\delta_1$$\uparrow$ & MAE(m)$\downarrow$ & AbsRel$\downarrow$ & $\delta_1$$\uparrow$ \\
\midrule
libfreenect2 & 0.0908 & 0.0659 & 0.9844 & 0.1143 & 0.0806 & 0.9541 \\
BM3D & 0.0982 & 0.0549 & 0.9733 & 0.1145 & 0.0799 & 0.9602 \\
ToFNet & 0.1425 & 0.2318 & 0.9346 & 0.1860 & 0.2155 & 0.8858 \\
UDA & 0.0470 & 0.0307 & 0.9916 & 0.0496 & 0.0337 & 0.9923 \\
RADU & 0.1197 & 0.0942 & 0.9423 & 0.0913 & 0.0845 & 0.9592 \\
Palette & 1.1692 & 1.1293 & 0.2274 & 1.5315 & 1.5679 & 0.1296 \\
DepthGen & 0.8851 & 0.7165 & 0.2544 & 0.8801 & 0.8306 & 0.2723 \\
\midrule
\textbf{Ours} & \textbf{0.0148} & \textbf{0.0297} & \textbf{0.9933} & \textbf{0.0163} & \textbf{0.0212} & \textbf{0.9959} \\
\bottomrule
\end{tabular}
}
\label{tab:comparison}
\vspace{-0.1cm}
\end{table}

\begin{table}[t]
\centering
\caption{Comparison of quantitative evaluation on FLAT testing dataset with DepthCAD variants}
\vspace{-0.2cm}
\resizebox{\columnwidth}{!}{ 
\begin{tabular}{cccccccc}
\toprule
\multicolumn{3}{c}{Modules} & \multicolumn{5}{c}{Metrics}  \\ 
\cmidrule(lr){1-3} \cmidrule(lr){4-8}
noisy raw & confidence & SD & MAE(m)$\downarrow$ & AbsRel$\downarrow$ & $\delta_1\uparrow$ & $\delta_2\uparrow$ & $\delta_3\uparrow$ \\ 
\midrule
- & - & SD2.1 & 0.8239 & 0.4791 & 0.7917 & 0.8554 & 0.8829 \\ 
- & \checkmark & SD2.1 & 0.7525 & 0.8915 & 0.2558 & 0.3671 & 0.5125 \\
HDR & \checkmark & SD2.1 & 0.0646 & 0.0883 & 0.9686 & 0.9733 & 0.9756 \\
LDR & - & SD2.1 & 0.0515 & 0.0342 & 0.9903 & 0.9912 & 0.9928 \\ 
LDR & \checkmark & SD1.5 & 0.0460 & 0.0969 & 0.9796 & 0.9933 & 0.9980 \\ 
\midrule
\multicolumn{3}{c}{\textbf{Ours}} & \textbf{0.0148} & \textbf{0.0297} & \textbf{0.9933} & \textbf{0.9975} & \textbf{0.9989} \\
\bottomrule
\end{tabular}
}
\label{tab:ablation}
\vspace{-0.3cm}
\end{table}

\section{Conclusion}
\label{sec:con}
In this work, we present DepthCAD, a novel ToF depth denoising approach based on diffusion. By performing diffusion on low-dynamic-range raw ToF correlations, our approach effectively mitigates the domain gap between color images and ToF data. Furthermore, our approach steers the diffusion process to produce high-fidelity depth reconstructions with noisy raw ToF correlation measurements and confidence as guidance. Experimental results on FLAT dataset and its augmented version demonstrate the superior denoising accuracy of our approach on synthetic data, as well as enhanced robustness to real-world noise over competing schemes.

\bibliographystyle{IEEEbib}
\bibliography{main}

\begin{thebibliography}{10}

\bibitem{zanuttigh2016time}
Pietro Zanuttigh, Giulio Marin, Carlo Dal~Mutto, Fabio Dominio, Ludovico Minto, Cortelazzo, et~al.,
\newblock ``Time-of-flight and structured light depth cameras,''
\newblock {\em Technology and Applications}, vol. 978, no. 3, 2016.

\bibitem{zhang2015fusion}
Yueyi Zhang, Zhiwei Xiong, and Feng Wu,
\newblock ``Fusion of time-of-flight and phase shifting for high-resolution and low-latency depth sensing,''
\newblock in {\em ICME}. IEEE, 2015, pp. 1--6.

\bibitem{zeng2019deep}
Jin Zeng, Yanfeng Tong, Yunmu Huang, Qiong Yan, Wenxiu Sun, Jing Chen, and Yongtian Wang,
\newblock ``Deep surface normal estimation with hierarchical rgb-d fusion,''
\newblock in {\em CVPR}, 2019, pp. 6153--6162.

\bibitem{gao2021joint}
Rongrong Gao, Na~Fan, Changlin Li, Wentao Liu, and Qifeng Chen,
\newblock ``Joint depth and normal estimation from real-world time-of-flight raw data,''
\newblock in {\em IROS}. IEEE, 2021, pp. 71--78.

\bibitem{kurillo2022evaluating}
Gregorij Kurillo, Evan Hemingway, Mu-Lin Cheng, and Louis Cheng,
\newblock ``Evaluating the accuracy of the {A}zure {K}inect and {K}inect v2,''
\newblock {\em Sensors}, vol. 22, no. 7, pp. 2469, 2022.

\bibitem{xiang2016libfreenect2}
Lingzhu Xiang, Florian Echtler, Christian Kerl, Thiemo Wiedemeyer, R~Gordon, and F~Facioni,
\newblock ``libfreenect2: Release 0.2,'' 2016.

\bibitem{chen2020very}
Yan Chen, Jimmy Ren, Xuanye Cheng, Keyuan Qian, Luyang Wang, and Jinwei Gu,
\newblock ``Very power efficient neural time-of-flight,''
\newblock in {\em WACV}, 2020, pp. 2257--2266.

\bibitem{deng2023two}
Yufan Deng, Xin Deng, and Mai Xu,
\newblock ``A two-stage hybrid cnn-transformer network for rgb guided indoor depth completion,''
\newblock in {\em ICME}. IEEE, 2023, pp. 1127--1132.

\bibitem{su2018deep}
Shuochen Su, Felix Heide, Gordon Wetzstein, and Wolfgang Heidrich,
\newblock ``Deep end-to-end time-of-flight imaging,''
\newblock in {\em CVPR}, 2018, pp. 6383--6392.

\bibitem{uda2019}
Gianluca Agresti, Henrik Schaefer, Piergiorgio Sartor, and Pietro Zanuttigh,
\newblock ``Unsupervised domain adaptation for tof data denoising with adversarial learning,''
\newblock in {\em CVPR}, 2019, pp. 5579--5586.

\bibitem{schelling2022radu}
Michael Schelling, Pedro Hermosilla, and Timo Ropinski,
\newblock ``Radu: Ray-aligned depth update convolutions for tof data denoising,''
\newblock in {\em CVPR}, 2022, pp. 671--680.

\bibitem{jia2025deep}
Jingwei Jia, Changyong He, Jianhui Wang, Gene Cheung, and Jin Zeng,
\newblock ``Deep unrolled graph laplacian regularization for robust time-of-flight depth denoising,''
\newblock {\em IEEE Signal Processing Letters}, 2025.

\bibitem{barron2016fast}
Jonathan~T Barron and Ben Poole,
\newblock ``The fast bilateral solver,''
\newblock in {\em European conference on computer vision}. Springer, 2016, pp. 617--632.

\bibitem{georgiev2018time}
Mihail Georgiev, Robert Bregovi{\'c}, and Atanas Gotchev,
\newblock ``Time-of-flight range measurement in low-sensing environment: Noise analysis and complex-domain non-local denoising,''
\newblock {\em IEEE TIP}, vol. 27, no. 6, pp. 2911--2926, 2018.

\bibitem{dabov2007image}
Kostadin Dabov, Alessandro Foi, Vladimir Katkovnik, and Karen Egiazarian,
\newblock ``Image denoising by sparse 3-d transform-domain collaborative filtering,''
\newblock {\em TIP}, vol. 16, no. 8, pp. 2080--2095, 2007.

\bibitem{hu2013depth}
Wei Hu, Xin Li, Gene Cheung, and Oscar Au,
\newblock ``Depth map denoising using graph-based transform and group sparsity,''
\newblock in {\em MMSP}. IEEE, 2013, pp. 001--006.

\bibitem{rossi2020joint}
Mattia Rossi, Mireille~El Gheche, et~al.,
\newblock ``Joint graph-based depth refinement and normal estimation,''
\newblock in {\em CVPR}, 2020, pp. 12154--12163.

\bibitem{dong2025exploiting}
Guanting Dong, Yueyi Zhang, Xiaoyan Sun, and Zhiwei Xiong,
\newblock ``Exploiting dual-correlation for multi-frame time-of-flight denoising,''
\newblock in {\em European Conference on Computer Vision}. Springer, 2024, pp. 473--489.

\bibitem{guo2018tackling}
Qi~Guo, Iuri Frosio, Orazio Gallo, et~al.,
\newblock ``Tackling 3d tof artifacts through learning and the flat dataset,''
\newblock in {\em ECCV}, 2018, pp. 368--383.

\bibitem{saharia2022palette}
Chitwan Saharia, William Chan, Huiwen Chang, Chris Lee, Jonathan Ho, Tim Salimans, David Fleet, and Mohammad Norouzi,
\newblock ``Palette: Image-to-image diffusion models,''
\newblock in {\em SIGGRAPH}, 2022, pp. 1--10.

\bibitem{ho2020denoising}
Jonathan Ho, Ajay Jain, and Pieter Abbeel,
\newblock ``Denoising diffusion probabilistic models,''
\newblock {\em NeurIPS}, vol. 33, pp. 6840--6851, 2020.

\bibitem{wang2022zero}
Yinhuai Wang, Jiwen Yu, and Jian Zhang,
\newblock ``Zero-shot image restoration using denoising diffusion null-space model,''
\newblock {\em ICLR}, 2023.

\bibitem{wang2024exploiting}
Jianyi Wang, Zongsheng Yue, et~al.,
\newblock ``Exploiting diffusion prior for real-world image super-resolution,''
\newblock {\em IJCV}, pp. 1--21, 2024.

\bibitem{frank2009theoretical}
Mario Frank, Matthias Plaue, Holger Rapp, Ullrich K{\"o}the, Bernd J{\"a}hne, and Fred~A Hamprecht,
\newblock ``Theoretical and experimental error analysis of continuous-wave time-of-flight range cameras,''
\newblock {\em Optical Engineering}, vol. 48, no. 1, pp. 013602--013602, 2009.

\bibitem{rombach2022high}
Robin Rombach, Andreas Blattmann, Dominik Lorenz, Patrick Esser, and Bj{\"o}rn Ommer,
\newblock ``High-resolution image synthesis with latent diffusion models,''
\newblock in {\em CVPR}, 2022, pp. 10684--10695.

\bibitem{saxena2023monocular}
Saurabh Saxena, Abhishek Kar, Mohammad Norouzi, and David~J. Fleet,
\newblock ``Monocular depth estimation using diffusion models,'' 2023.

\bibitem{paszke2017automatic}
Adam Paszke, Sam Gross, Soumith Chintala, Gregory Chanan, Edward Yang, Zachary DeVito, Zeming Lin, Alban Desmaison, Luca Antiga, and Adam Lerer,
\newblock ``Automatic differentiation in pytorch,''
\newblock 2017.

\bibitem{song2020denoising}
Jiaming Song, Chenlin Meng, and Stefano Ermon,
\newblock ``Denoising diffusion implicit models,''
\newblock {\em arXiv preprint arXiv:2010.02502}, 2020.

\bibitem{ke2024repurposing}
Bingxin Ke, Anton Obukhov, Shengyu Huang, Nando Metzger, Rodrigo~Caye Daudt, et~al.,
\newblock ``Repurposing diffusion-based image generators for monocular depth estimation,''
\newblock in {\em CVPR}, 2024, pp. 9492--9502.

\end{thebibliography}

\end{document}